\documentclass[10pt,twocolumn,letterpaper]{article}

\usepackage{cvpr}              

\usepackage{bbm}
\usepackage{overpic}
\usepackage{graphicx}
\usepackage{amsmath}
\usepackage{amssymb}
\usepackage{booktabs}
\usepackage{multirow}
\usepackage[table]{xcolor}

\usepackage{enumitem}
\usepackage{pifont}
\usepackage{comment}
\newcommand{\cmark}{\ding{51}}
\newcommand{\xmark}{\ding{55}}

\usepackage{mathtools} 
\usepackage{cuted} 

\usepackage{epigraph}

\usepackage[pagebackref,breaklinks,colorlinks]{hyperref}

\usepackage[capitalize]{cleveref}
\crefname{section}{Sec.}{Secs.}
\Crefname{section}{Section}{Sections}
\Crefname{table}{Table}{Tables}
\crefname{table}{Tab.}{Tabs.}

\def\cvprPaperID{4313} 
\def\confName{CVPR}
\def\confYear{2022}

\begin{document}
\title{DynamicEarthNet: Daily Multi-Spectral Satellite Dataset \\for Semantic Change Segmentation}
\author{Aysim Toker$^{1,*}$, Lukas Kondmann$^{1,2,*}$, Mark Weber$^{1}$, Marvin Eisenberger$^{1}$, Andrés Camero$^{2}$, \\
Jingliang Hu$^{2}$, Ariadna Pregel Hoderlein$^{1}$, Çağlar Şenaras$^{3}$, Timothy Davis$^{3}$, \\
Daniel Cremers$^{1}$,Giovanni Marchisio$^{3,\dagger}$, Xiao Xiang Zhu$^{1,2,\dagger,\ddagger}$, Laura Leal-Taixé$^{1,\dagger}$\\[5pt]
Technical University of Munich$^{1}$, German Aerospace Center$^{2}$, Planet Labs$^{3}$\\
}
\newcommand{\parag}[1]{\vspace{0ex} \textit{#1}}

\newcommand{\PAR}[1]{\vskip4pt \noindent {\bf #1~}}
\newcommand{\CAP}[1]{{\bf #1~}}
\newcommand{\PARbegin}[1]{\noindent {\bf #1~}}
\newcommand{\TODO}[1]{\textcolor{red}{#1}}
\newcommand{\DONE}[1]{\textcolor{green}{#1}}
\newcommand{\REF}{\textcolor{red}{\textbf{REF}}}
\newcommand{\changed}[1]{\textcolor{red}{#1}}

\newcommand{\mar}[1]{\textcolor{orange}{\textbf{Mark: }{#1}}}
\newcommand{\aysim}[1]{\textcolor{purple}{\textbf{Aysim: }{#1}}}
\newcommand{\lukas}[1]{\textcolor{blue}{\textbf{Lukas: }{#1}}}
\newcommand{\marvin}[1]{\textcolor{red}{\textbf{Marvin: }{#1}}}
\newcommand{\lau}[1]{\textcolor{magenta}{\textbf{Laura: }{#1}}}

\setlength{\floatsep}{5pt plus2pt minus4pt}
\setlength{\textfloatsep}{5pt plus2pt minus4pt}
\setlength{\dblfloatsep}{5pt plus2pt minus4pt}
\setlength{\dbltextfloatsep}{5pt plus2pt minus4pt}

\newcommand{\metric}{SCS }
\newcommand{\metricfull}{Semantic Change Segmentation (SCS) }
\newcommand{\bmetric}{BC}
\newcommand{\smetric}{SC}

\newcommand{\dynnet}{\textit{DynamicEarthNet}}
\newcommand{\tx}{\mathbf{x}}
\newcommand{\ty}{\mathbf{y}}
\newcommand{\tb}{\mathbf{b}}
\newcommand{\tyh}{\hat{\mathbf{y}}}
\newcommand{\tbh}{\hat{\mathbf{b}}}
\newcommand{\tm}{\mathbf{m}}
\newcommand{\iou}{\mathrm{IoU}}
\newcommand{\bbone}{\mathbbm{1}}
\newcommand{\cC}{\mathcal{C}}
\twocolumn[{%
\renewcommand\twocolumn[1][]{#1}%
\vspace{-4.5em}
\maketitle
\thispagestyle{empty}
\begin{center}
    \vspace{-2em}
    \centering
    \includegraphics[width=1.\textwidth]{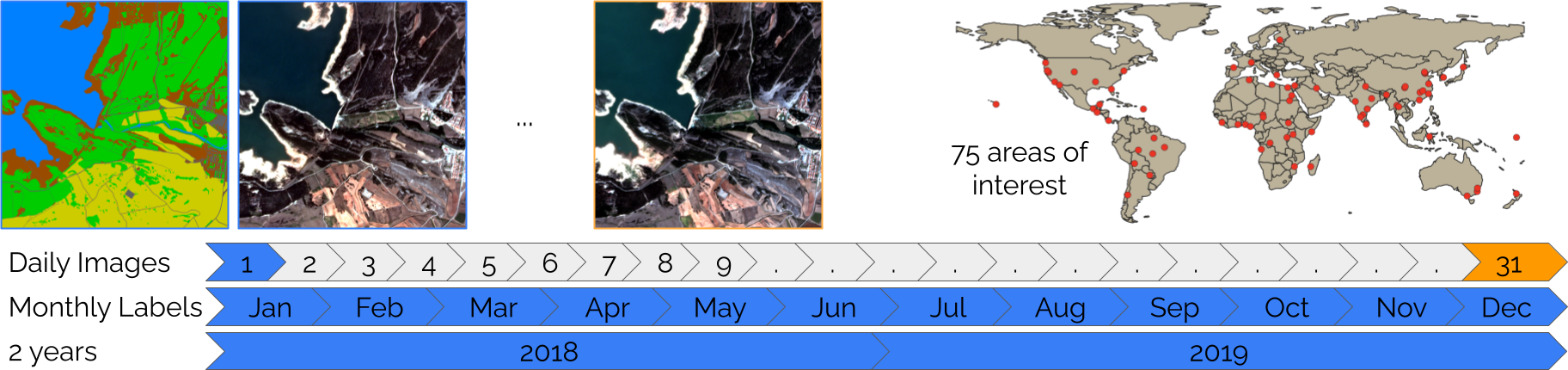}
    \captionof{figure}{\textbf{Visualization of the \dynnet\ dataset.}
    For a specific area of interest, we show two satellite observations, 2019-08-01 and 2019-08-31, as well as the corresponding monthly ground-truth annotation (top left). The complete dataset consists of daily samples in the range from 2018-01-01 to 2019-12-31.
    We consider 75 separate areas of interest, spread over six continents (top right).}
    \label{fig:teaser}
\end{center}
}]

\begin{abstract}
Earth observation is a fundamental tool for monitoring the evolution of land use in specific areas of interest. Observing and precisely defining change, in this context, requires both time-series data and pixel-wise segmentations. To that end, we propose the \dynnet\ dataset that consists of daily, multi-spectral satellite observations of 75 selected areas of interest distributed over the globe with imagery from Planet Labs. These observations are paired with pixel-wise monthly semantic segmentation labels of 7 land use and land cover (LULC) classes. \dynnet\ is the first dataset that provides this unique combination of daily measurements and high-quality labels. In our experiments, we compare several established baselines that either utilize the daily observations as additional training data (semi-supervised learning) or multiple observations at once (spatio-temporal learning) as a point of reference for future research. Finally, we propose a new evaluation metric \metric that addresses the specific challenges associated with time-series semantic change segmentation. The data is available at: \url{https://mediatum.ub.tum.de/1650201}.

\epigraph{Making peace with nature is the defining task of the 21st century.}{António Guterres, UN Secretary General}
\vspace{-2em}
\makeatletter{\renewcommand*{\@makefnmark}{}
\footnotetext{* Authors share first authorship. $\dagger$ Authors share senior authorship. $\ddagger$ Corresponding author: xiaoxiang.zhu@dlr.de.}\makeatother}
\end{abstract}
\section{Introduction}
\label{sec:intro}
Society is rapidly becoming more aware of the human footprint on the world's climate. Overwhelming evidence shows that climate change has both short-term and long-term effects on almost every aspect of our lives~\cite{ipcc2021}. Using simulations and global climate metrics, it is nowadays possible to observe changes at a global scale, like the rising sea levels or changes of the gulf stream. In contrast, precise predictions of local changes are much harder to obtain. Common examples include land use by agriculture, deforestation, flooding, wildfires, growth of urban areas, and transportation infrastructure. It is of critical importance to monitor such local changes since these are the factors that ultimately exacerbate the global climate crisis. 

Satellite images are a powerful tool in this context to track local changes to the environment in specific regions. Observing change at a local scale requires two conditions: high frequency of satellite observations and pixel-precise understanding of the observed surface. Existing datasets often fail to provide these conditions. Whenever pixel-wise annotations are provided, only static images can be used~\cite{waqas2019cvprw} or the revisit frequency is limited to once a year~\cite{daudt2019multitask, tian2020neuripsw}. Datasets with coarser annotations have either an irregular~\cite{christie2018cvpr} or monthly revisit frequency~\cite{van2021cvpr}. As an example of land changes, in 2020, 46${km}^2$ of the rainforest in Brazil were destroyed every day~\cite{gfw2021}. This suggests that if we analyze the satellite images of that area once per month, we potentially miss deforestation of the equivalent of the city of Los Angeles, California. As Brazil alone has millions of square kilometers of forest, automatic methods are required to detect these and other kinds of land changes. 
Current pixel-precise automatic methods are predominantly based on deep learning and thus require annotated data to learn.

In this work, we present \dynnet, a time-series satellite imagery dataset with daily revisits of 75 local regions across the globe. The dataset comprises consistent, occlusion-free daily observations with multi-spectral imagery over the span of two years (2018-2019). We further provide annotated monthly semantic segmentation labels. The main focus is to segment and detect changes in the development of general land use and land cover (LULC). Specifically, we focus on the following LULC classes: impervious surfaces, water, soil, agriculture, wetlands, snow \& ice, and forest \& other vegetation.

In comparison to semantic segmentation on standard computer vision benchmarks, satellite imagery is subject to various additional challenges. Most prominently, labeled areas in satellite images typically have very intricate shapes that are significantly more complex than everyday objects. We show that well-performing methods~\cite{ronneberger2015u,chen2018eccv} on standard vision benchmarks do not necessarily transfer well to this domain. Furthermore, common segmentation metrics are not optimal for quantifying the performance on the task of semantic change segmentation. We alleviate this issue by proposing a new evaluation protocol that captures the essence of semantic change segmentation. \dynnet \ and the proposed evaluation protocol encourage the development of more specialized algorithms that can handle the particular challenges of daily time-series satellite imagery. In summary, our contributions are as follows:
\begin{itemize}
    \item We present a large-scale dataset of multi-spectral satellite imagery with daily observations of 75 separate areas of interest around the globe.
    \item We provide dense, monthly annotations of 7 land use and land cover (LULC) semantic classes.
    \item We propose a novel evaluation protocol that models two central properties of semantic change segmentation: binary change and semantic segmentation. 
    \item We evaluate multiple baseline approaches on our data for the task of detecting semantic change. We show how the time-series nature of our data can be leveraged for optimal performance.
\end{itemize}

\section{Related work}
\label{sec:related}
For our discussion of related work, we provide an overview of publicly available satellite imagery datasets, see also~\cref{table:relatedsatellites}. Furthermore, we summarize existing work on the tasks of semantic segmentation and change detection.

\begin{table*}
\begin{center}
\scalebox{0.80}{
\begin{tabular}{lccrcccc}
\toprule[0.2em]
Dataset & Temporal & Revisit Time & \# Images & Sources & GSD (m) &  Annotation & Objects\\
\toprule[0.2em]
SpaceNet~\cite{van2018arxiv}    & \xmark & \xmark  &   $>$24,586 & Maxar & 0.31 & Polygon & Buildings and Roads\\
DOTA~\cite{xia2018cvpr}        & \xmark & \xmark  & 2,806 & Google Earth & 0.15-12$^\ddagger$ & Oriented Bbox & Various \\
fMoW~\cite{christie2018cvpr}    & \cmark & irregular & $>$1,000,000 & Maxar & 0.31-1.60 & BBox & Various\\
SpaceNet MVOI~\cite{weir2019iccv} & \xmark & \xmark & 60,000 & Maxar & 0.46-1.67 & Polygon & Buildings\\
MUDS~\cite{van2021cvpr}        & \cmark &  monthly & 2,389 & Planet & 4.0  & Polygon & Buildings \\ 
DOTA-v2.0~\cite{ding2021tpami} & \xmark & \xmark & 11,268 & Google Earth &  0.15-12$^\ddagger$ & Oriented Bbox & Various \\
\hline
DeepGlobe~\cite{demir2018cvprw} & \cellcolor[HTML]{DAE8FC}\xmark & \cellcolor[HTML]{FFFFB3}\xmark  & 1,146   & Maxar & 0.5  & Seg. Mask & Various LULC\\
iSAID~\cite{waqas2019cvprw}     & \cellcolor[HTML]{DAE8FC}\xmark & \cellcolor[HTML]{FFFFB3}\xmark  & 2,806 & DOTA & 0.15-12$^\ddagger$ & I. Seg Mask & Various\\
HRSCD~\cite{daudt2019multitask}  & \cellcolor[HTML]{DAE8FC}\cmark & \cellcolor[HTML]{FFFFB3}yearly  & 582  & BD ORTHO & 0.5 & Seg. Mask & Various LULC \\
Hi-UCD~\cite{tian2020neuripsw}  & \cellcolor[HTML]{DAE8FC}\cmark & \cellcolor[HTML]{FFFFB3}yearly  & 2,586  & ELB$^\dagger$ & 0.1 & Seg. Mask & Various LULC \\
\hline
\textbf{\dynnet}        & \cellcolor[HTML]{DAE8FC}\cmark & \cellcolor[HTML]{FFFFB3}daily & 54,750 & PlanetFusion & 3.0 & Seg. Mask& Various LULC\\ \bottomrule[0.1em]
\end{tabular}
}
\\{\qquad\scriptsize$^\dagger$ Estonian Land Board, $^\ddagger$ Google Earth gathers information from various sensors, so the resolution is diverse~\cite{weir2019iccv}.}
\centering
    \caption{\textbf{An overview of public satellite datasets.} For each dataset, we compare key characteristics like the revisit time, the number of images, data source, ground sample distance (GSD), types of annotations, and annotated objects. Most closely related are DeepGlobe~\cite{demir2018cvprw}, iSAID~\cite{waqas2019cvprw}, HRSCD~\cite{daudt2019multitask} and Hi-UCD~\cite{tian2020neuripsw} which, like ours, provide dense semantic annotations for various land cover classes. However, they either provide no time-series data or merely yearly revisit times. Closely related datasets are highlighted in blue and yellow.}
    \label{table:relatedsatellites}
\end{center}
  \footnotetext[1]{table footnote 1} 

\end{table*}

\subsection{Earth observation datasets}\label{subsec:relateddatasets}
\PAR{Segmentation and detection.}
Semantic segmentation of land cover classes for satellite imagery was originally pioneered by the ISPRS project~\cite{isprs20182vaihingen,isprs20182potsdam}. Similarly, the DeepGlobe~\cite{demir2018cvprw} and SpaceNet~\cite{van2018arxiv} challenges provide datasets for building detection, road extraction, and land cover classification. In contrast to ours, such early works have a relatively small number of areas of interest.

Subsequently, the main focus started to shift towards large-scale aerial imagery~\cite{xia2018cvpr,waqas2019cvprw}. To that end, DOTA~\cite{xia2018cvpr} proposes to detect objects on a large collection of images cropped from Google Earth. iSAID~\cite{waqas2019cvprw} extends this concept to the task of instance segmentation. Along the same lines, SpaceNet MVOI~\cite{weir2019iccv} proposes a benchmark on building detection for multi-view satellite imagery. Our benchmark, on the other hand, provides semantic annotations that are dense, \ie defined for every single pixel.
\PAR{Change detection.}
Several works aim at predicting change between observations of the same area of interest at different times. Most relevant datasets focus on binary change detection which is agnostic to specific types of change~\cite{bourdis2011ieeegrs,daudt2018igarss}. HRSCD~\cite{daudt2019multitask} and Hi-UCD~\cite{tian2020neuripsw} propose a multi-class semantic change detection datasets. In comparison to time-series data, these benchmarks show only one observation per year, for 2-3 years in total, rather than a full sequence. Moreover, the diversity is limited -- HRSCD~\cite{daudt2019multitask} and Hi-UCD~\cite{tian2020neuripsw} cover specific regions of France and Tallinn, Estonia, respectively. More recently, QFabric~\cite{verma2021qfabric} presented a large-scale multi-temporal dataset, with polygonal annotations for change regions. In contrast, our dataset contains daily observations and pixel-wise LULC classes.

\PAR{Time-series analysis.}
In recent times, time-series satellite datasets gained increasing attention~\cite{christie2018cvpr,van2021cvpr,requena2021earthnet2021}. For instance, Earthnet2021~\cite{requena2021earthnet2021} presents a surface forecasting dataset based on public Sentinel-2 imagery with a revisit rate of 5 days. Since the intended applications are quite dissimilar to ours, no land cover annotations are provided. fMoW~\cite{christie2018cvpr} provides temporal satellite imagery with bounding box annotations. Similarly, MUDS~\cite{van2021cvpr} aims at monitoring urbanization by tracking buildings for several areas of interest that are annotated with polygons. Varying acquisition conditions make it challenging to consistently collect data over an extended period of time. Consequently, existing datasets often contain irregular revisit frequencies~\cite{christie2018cvpr} or infrequent (monthly) observation intervals~\cite{van2021cvpr}. In contrast, our \dynnet \ dataset provides high-quality, consistent daily observations. 

\subsection{Considered tasks}
\label{subsec:consideredtasks}
\PAR{Semantic segmentation.}
There are countless recent deep learning methods~\cite{ronneberger2015u,long2015cvpr,chen2017deeplab,badrinarayanan2017tpami,chen2017rethinking,chen2018eccv,wang2020tpami} that address general semantic segmentation.
In comparison to most common computer vision applications, segmentation of satellite images is subject to specific challenges, such as irregular sizes and shapes of segmented regions. Recent approaches show that encoder-decoder architectures~\cite{lin2017cvpr,kirillov2019cvpr} can help to address the foreground-background imbalance of satellite data~\cite{zheng2020cvpr,li2021cvpr}. Most existing algorithms focus on segmenting individual, static images. A few works leverage the additional information from time-series satellite images for the case of crop-type classification~\cite{rustowicz2019cvprw,garnot2021iccv,kondmann2021denethor}. 
We believe that the \dynnet \ dataset will encourage researchers to develop specialized algorithms that can handle the particular challenges of time-series satellite imagery.
\PAR{Change detection.} Change detection is an extensively studied topic in earth observation. Classical approaches define axiomatic, pixel-based~\cite{bovolo2006ieeetrans, bovolo2008ieeetrans,bovolo2011ieeetrans,thonfeld2016robust,kondmann2021arxiv} algorithms to obtain change whereas many recent approaches are data-driven~\cite{zhan2017change,daudt2018icip,chen2019arxiv,saha2019ieeetransactions}. The development of new algorithms is often inhibited by a lack of high-quality data and expert annotations. Most methods focus on binary change and are usually limited to two distinct observations in time (bitemporal)~\cite{bovolo2006ieeetrans, bovolo2008ieeetrans,bovolo2011ieeetrans,thonfeld2016robust,kondmann2021arxiv,zhan2017change,chen2019arxiv}. Moreover, datasets and metrics used for evaluation differ widely and are often not public. 

These considerations underline the necessity for a standardized benchmark with a consistent evaluation protocol. Up to now, there are few approaches suitable for multi-class change detection. Most of them typically consider two snapshots, often years apart. Among these works,~\cite{lyu2016remotesensing,mou2019spectral} directly predict the multi-class change map whereas,~\cite{tian2020neuripsw} define change as the difference between two semantic maps. We follow the latter approach in our evaluations since existing work on multi-class change detection is not primarily designed to handle high temporal frequencies. Therefore, we benchmark state-of-the-art semantic segmentation algorithms on our dataset and compare differences in the predicted multi-class semantic masks over time.
 \begin{figure*}
    \centering
    \includegraphics[width=1.0\linewidth]{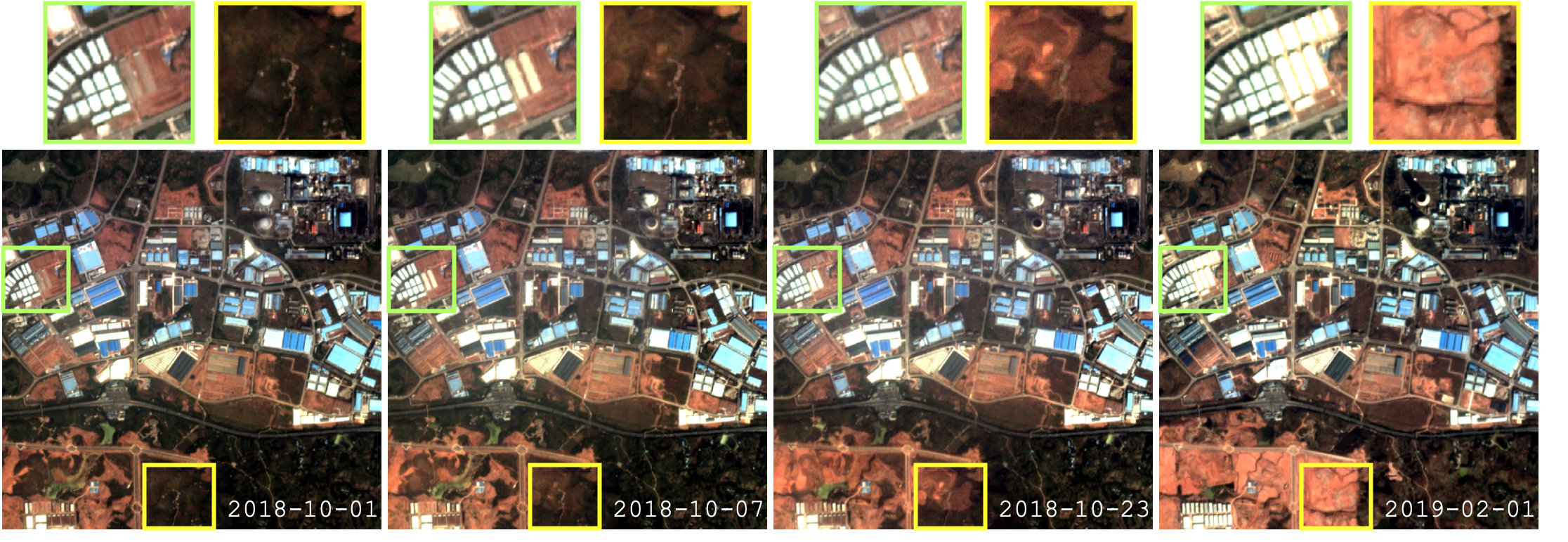}
    \caption{\textbf{An example of a changing surface.} We show four sample frames of one AOI from our dataset at different times. Two sub-regions are magnified that highlight two types of change we encounter in practice (top row). The daily nature of our data allows us to observe new buildings being built (green) or to track deforestation (yellow). Additionally, we can monitor the long-term effects of such changes over the span of multiple months, \eg the changes to the forest patch here are persistent.} 
    \label{fig:dataset_sequence}
\end{figure*}

\section{The \dynnet \ dataset}
\label{sec:dataset}
We present the \dynnet \ dataset that contains daily, cloud-free satellite data acquired from January 2018 to December 2019. 
It consists of images from 75 areas of interest (AOIs) across the globe, as illustrated by the world map in~\cref{fig:teaser}. The dataset covers a wide variety of environments with diverse types of land cover changes.  
For each region, we provide a sequence of images with daily revisits.
Furthermore, we present pixel-wise semantic labels for the first day of each month.
These serve as ground-truth to define land cover changes over the span of two observed years.
In the remainder of this section, we provide details on the imagery, semantic labels, and statistics of the dataset.
\begin{table}[]
\vspace{0.5em}
\begin{center}
\begin{tabular}{llcl}
\toprule[0.2em]
class name                 & \%   & \#AOIs & color   \\
\toprule[0.2em]
impervious surface         & 7.1  & 70 &\cellcolor[HTML]{606060}\\
agriculture                & 10.3 & 37 &\cellcolor[HTML]{cccc00}\\
forest \& other vegetation & 44.9 & 71 &\cellcolor[HTML]{00cc00}\\
wetlands                   & 0.7  & 24 &\cellcolor[HTML]{000099}\\
soil                       & 28.0 & 75 &\cellcolor[HTML]{994c00}\\
water                      & 8.0  & 58 &\cellcolor[HTML]{0080ff}\\
snow \& ice                & 1.0  & 2 &\cellcolor[HTML]{8ab2c6}
\end{tabular}
\centering
    \caption{\textbf{LULC class distribution.} The distribution of LULC classes averaged over all $24\times 75=1800$ semantic maps in the dataset. Additionally, we report the absolute number of AOIs with any occurrences of a given LULC class. We visualize the colors we use for each class throughout the paper.}
    \label{table:pixdistribution}
\end{center}
\end{table}
\subsection{Multi-spectral imagery}\label{subsec:images}
The primary source of our dataset is the Fusion Monitoring product\footnote{https://www.planet.com/pulse/planet-announces-powerful-new-products-at-planet-explore-2020/} from Planet Labs, which provides multi-spectral time-series satellite imagery. Each snapshot contains four channels (RGB + near-infrared) with a ground sample distance (GSD), \ie pixel granularity, of 3 meters and a resolution of 1024x1024. 

Beyond the raw observational data, Planet applies a combination of post-processing techniques to ensure data quality and consistency: For once, all images are processed to remove occlusions by weather, overcast and related visual artifacts.
The data is gap-filled, which means that missing information due to cloud coverage is filled with suitable observations from the closest available point in time. Moreover, the Fusion bands are calibrated to the Harmonized Landsat-Sentinel (HLS)\footnote{https://earthdata.nasa.gov/esds/harmonized-landsat-sentinel-2} spectrum to make them compatible with other publicly available datasets such as Landsat 8~\cite{woodcock2008free} or Sentinel 2~\cite{drusch2012sentinel,aschbacher2017esa}. 

To encourage the exploration of data fusion, we provide monthly Sentinel-2 (S2) imagery from the same 75 AOIs for reference. The main idea of this auxiliary set of images is to allow for comparisons with publicly available data. Moreover, the additional data potentially gives rise to interesting multi-modal settings in future experiments. For more details, we refer the reader to our supplementary material.

\subsection{Pixel-wise labels}\label{subsec:pixlabels}
Having described the raw satellite imagery, we now provide more details on the monthly ground-truth annotations. They comprise a collection of pixel-wise semantic segmentation labels corresponding to the first day of each month. These labels are defined as the common LULC classes, \ie, impervious surfaces, agriculture, forest \& other vegetation, wetlands, soil, water, snow \& ice. The resolution of each annotation is 1024x1024 with a pixel granularity of 3 meters, just like the corresponding satellite images.

The annotation procedure was rigorous with an emphasis on the temporal consistency of the labels. The first image was manually annotated for each AOI and used as a basis for the following months. Subsequent maps are updated if there is a perceptible change in a certain region that is evident to the human annotator. Three quality control gates, each with a different annotator, ensure accurate annotations, topological correctness, and format correctness, respectively.

\subsection{Dataset statistics}\label{subsec:dataprep}
The \dynnet \ dataset contains 75 different AOIs across the globe, each of which consists of a sequence of 730 images covering two years from January 2018 to December 2019. We provide semantic LULC classes for the first day of each month, 24 per sequence in total. In total, this amounts to $54750$ satellite images and $1800$ ground-truth annotations.

We illustrate the distribution of LULC classes over the whole dataset in~\cref{table:pixdistribution}. Due to the nature of the data, occurrences of certain semantic classes are imbalanced with forest \& other vegetation and soil dominating less frequent classes like wetlands. Such general ambient classes often take up large portions of a considered region, see the bottom third of the images in~\cref{fig:dataset_sequence}.

We split our data into train, validation, and test sets with 55, 10, and 10 AOIs, respectively. The number of distinct classes per AOI ranges from 2 to 6. For instance, some AOIs from the dataset contain only forest \& other vegetation and soil, whereas others include impervious surfaces, water, soil, agriculture, wetlands, and forest \& other vegetation. No single AOI contains all 7 classes. For an optimal balance, we ensure that the splits' classes are distributed as equally as possible. We refrain from providing more fine-grained statistics on the class distribution to avoid disclosing any additional information on the (currently concealed) test set. Since the snow \& ice class occurs in only 2 cubes, see ~\cref{table:pixdistribution}, we have no such examples in the validation or test sets. Consequently, we also do not consider this class in our quantitative evaluations presented in \cref{sec:baselines}.

\subsection{Advantages over existing benchmarks}

In comparison to other publicly available, annotated satellite datasets, \emph{DynamicEarthNet} has a number of crucial distinguishing features, see \cref{table:relatedsatellites}. First and foremost, it is the first to provide daily observations from a large diversity of AOIs. The closest work to ours in terms of revisit rates is~\cite{van2021cvpr} with monthly observations. Yet, they have a narrower focus with the main objective of tracking buildings to monitor urbanization. Other related change detection datasets~\cite{daudt2019multitask,tian2020neuripsw,verma2021qfabric} show merely one observation per year, see~\cref{table:relatedsatellites}. In our dataset, we provide consistent daily observations for two years allowing the study of both short-term and long-term change.~\cref{fig:dataset_sequence} highlights the potential of such data: We can observe the change of new buildings being built day by day. At the same time, we can pin down exact dates of deforestation, and successively observe long-term effects over the span of multiple months.
\section{Semantic change segmentation}
\label{sec:change}
One key application of our dataset is to measure how a given local region changes over time. For the standard task of binary change detection, we classify each pixel into change or no-change. This definition, however, disregards semantic information. We, therefore, generalize this classical notion to a multi-class segmentation task, which we refer to as semantic change segmentation.

For time-series satellite data, changes are usually caused by external forces, such as weather and climate effects or human destruction and creation. Compared to standard vision benchmarks, they often appear gradually over time and with a limited spatial extent. 
When predicting semantic labels for a whole observed region, such rare changes between frames have a low influence on the overall segmentation score. 
In our dataset, only 5\% of all pixels change from month to month on average. Hence, standard evaluation metrics defined on the full image like the Jaccard index ($\iou$) are not suitable to express how accurately semantic classes of changed areas are predicted.
We, therefore, propose a new metric to quantify the performance of methods in semantic change segmentation of satellite images. 

\subsection{Problem definition}\label{subsec:problemdef}
Let $\tx\in\mathbb{R}^{T\times H\times W\times 4}$ be an input time-series of satellite images consisting of $T$ frames with a spatial size of $H\times W$ and 4 input channels (RGB + near-infrared). For each such time-series, we further provide semantic annotations $\ty\in\cC^{T\times H\times W}$ that assign each pixel in $\tx$ to one of the 7 LULC classes $\cC:=\{0,\dots,6\}$ defined in \cref{subsec:pixlabels}. 
Given two consecutive frames at times $t$ and $t+1$, we can define the binary change $\tb\in\{0,1\}^{(T-1)\times H\times W}$ as a binary labeling of all pixels for which the ground-truth semantic label changes:
\begin{equation}\label{eq:binarychange}
    \tb_{t,i,j}:=\begin{cases}
    1, & \text{if}\quad\ty_{t,i,j}\neq\ty_{t-1,i,j},\\
    0, & \text{else}.
    \end{cases}
\end{equation}
When evaluating semantic change segmentation, both the binary change map $\tbh$ and the semantic map $\tyh$ need to be predicted. This requires methods to answer which pixels change and what class do these pixels change to.

\subsection{Evaluation protocol} 
There are two distinct types of errors that are common in the context of semantic change segmentation:
failing to detect the binary change and predicting the wrong semantic class for a changed pixel. Our goal is to design an evaluation protocol that captures both of these errors in a single signal.
Thus, the resulting semantic change segmentation (SCS) metric consists of two components, a class-agnostic binary change score (\bmetric) and a semantic segmentation score among changed pixels  (\smetric). 

\PAR{Binary change (\bmetric).} The standard approach to measure the quality of a predicted change map $\tbh$ is comparing its overlap with the ground-truth change $\tb$. This is commonly defined as the Jaccard index or intersection-over-union score
\begin{equation}\label{eq:bcscore}
    \mathrm{\bmetric}(\tb,\tbh)=\frac{|\{\tb=1\}\cap\{\tbh=1\}|}{|\{\tb=1\}\cup\{\tbh=1\}|}
\end{equation}

where we use the short hand-notation
\begin{equation}
\{\tb=1\}:=\{(t,i,j)\ |\ \tb_{t,i,j}=1\}
\end{equation}
for the indicator set of indices with binary change.

\PAR{Semantic change (\smetric).} The second component of our metric measures semantic change accuracy. It is defined as the segmentation score, conditioned on the set of pixels where any change occurs in the ground-truth maps, \ie $\tb=1$. On this subset of pixels, we compute the Jaccard index between the ground-truth labels $\ty$ and predicted labels $\tyh$ (averaged over all classes $c$):
\begin{equation}\label{eq:scscore}
    \mathrm{\smetric}(\ty,\tyh|\tb)=\frac{1}{|\cC|}\sum_{c\in\cC}\frac{\bigl|\{\tb=1\}\cap(\{\ty=c\}\cap\{\tyh=c\})\bigr|}{\bigl|\{\tb=1\}\cap(\{\ty=c\}\cup\{\tyh=c\})\bigr|}.
\end{equation}
\PAR{Semantic change segmentation (SCS).} The total \metric score is the arithmetic mean of the binary change and the semantic change:
\begin{equation}
    \mathrm{\metric}(\ty,\tyh) = \frac{1}{2}\bigl(\mathrm{BC}(\tb,\tbh) + \mathrm{SC}(\ty,\tyh|\tb)\bigr).
\end{equation}
In practice, we first accumulate confusion matrices of all
time-series before computing the final SCS score.
\PAR{Metric properties.}
In the following, we summarize a few distinguishing features of the proposed \metric metric. 
\setlist{nolistsep}
\begin{itemize}[itemsep=0pt]
    \item[\bf{i.}] {\bf Focus on change.} In comparison to standard metrics, like the Jaccard index, the SCS metric specifically emphasizes accurate change predictions.
    \item[\bf{ii.}] {\bf Separation of errors.} It separates the problems of detecting areas where change occurs (BC) and predicting the correct semantic labels for changed areas (SC).
    \item[\bf{iii.}] {\bf Single output signal.} Both signals contribute equally to the final SCS score.
\end{itemize}
\begin{table*}[t]
\begin{center}
\scalebox{0.90}{
\begin{tabular}{lcccccccc|c}
\toprule[0.2em]
 & Sample &  \multicolumn{6}{c}{\emph{per class IoU} ($\uparrow$)} &  Val & Test \\ \cline{3-8}
 & Frequency &  Imp. Surface & Agriculture & Forest & Wetlands & Soil & Water & mIoU ($\uparrow$)& mIoU ($\uparrow$)\\ \toprule[0.2em]
U-Net~\cite{ronneberger2015u} & monthly & 28.6 & 6.9 & 76.4 & 0.0 & 38.4 & 50.5 & 33.5 & 37.6 \\ \hline \hline
 \multirow{2}{*}{U-TAE~\cite{garnot2021iccv}}  & weekly & 31.8 & \textbf{8.0} & 77.3 & 0.0 & \textbf{39.1} & 58.1 & \textbf{35.7} & \textbf{39.7} \\
  & daily &     26.3             &  6.5    &  73.7    & 0.0 & 35.7   &  51.2    &  32.2  &  36.1   \\ \hline
  \multirow{2}{*}{U-ConvLSTM~\cite{rustowicz2019cvprw}}   & weekly  & 31.4    & 2.2      & \textbf{77.7}     &   0.0   &  36.1    &     58.6 &     {34.3} &         {39.1}\\
  &  daily &    14.4             &  0.6   &   72.1   &  0.0   &   32.0   &  58.8   &   29.7 &  30.9  \\ \hline
 \multirow{2}{*}{3D-Unet~\cite{rustowicz2019cvprw}} & weekly  & \textbf{32.4}     &    2.1  &  77.4    &  0.0    & 35.3     &   65.5  &  {35.5}  &  {37.2} \\ 
 &    daily &      31.1            &  1.8   &    75.8  &    0.0  &    34.1  &   \textbf{66.0}   &    34.8   &  38.8 \\ 
\bottomrule[0.1em]
\end{tabular}
}
\centering
\caption{\textbf{Quantitative results of spatio-temporal methods.} 
We compare the performance of different spatio-temporal architectures on the task of LULC segmentation. Individual values denote the intersection-over-union score for individual classes (cols. 3-8), as well as the averaged scores over the whole validation set (9th col.) and test set (10th col.). The monthly U-Net baseline is generally less accurate than the considered temporal architectures.
}
\label{tab:temporalbaselines}
\end{center}
\end{table*}
\begin{table*}
\begin{center}
\scalebox{0.90}{

\begin{tabular}{llcccccccc|c}
\toprule[0.2em]
&  & All & \multicolumn{6}{c}{\emph{per class IoU} ($\uparrow$)} & Val & Test \\ \cline{4-9}
&  & labelled? & Imp. Surface & Agriculture & Forest & Wetlands & Soil & Water & mIoU ($\uparrow$)& mIoU ($\uparrow$)\\
\toprule[0.2em]

\multirow{3}{*}{CAC~\cite{lai2021cvpr}}
& monthly      & \cmark & 18.1    &   4.8   &   74.7   &    0.0  &  33.9    &   \textbf{55.9}   &        31.2   &    37.9 \\
& weekly   & \xmark & 28.0  & \textbf{7.2}   & \textbf{75.7}    & \textbf{8.3}  & 38.9    & 51.0    &  \textbf{34.9}  &  37.9 \\
& daily    &  \xmark & \textbf{28.9}    &  4.0    &  75.5   &    0.5  & \textbf{39.0}     & 55.6    &    33.9   &       \textbf{43.6}             \\ 

\bottomrule[0.1em]
\end{tabular}
}
\centering
    \caption{\textbf{Quantitative results of semi-supervised methods.} The table shows the semantic segmentation results of using the context-aware consistency-based semi-supervised approach~\cite{lai2021cvpr} on our \dynnet\ dataset. We further present the IoU scores per class for the validation set. `Monthly' indicates that the architecture is trained in a supervised manner. Using unlabelled satellite images improves the results over the fully supervised baseline.}
    \label{tab:semisupervised}
\end{center}

\end{table*}

\begin{figure*}
    \centering
    \includegraphics[width=1.0\linewidth]{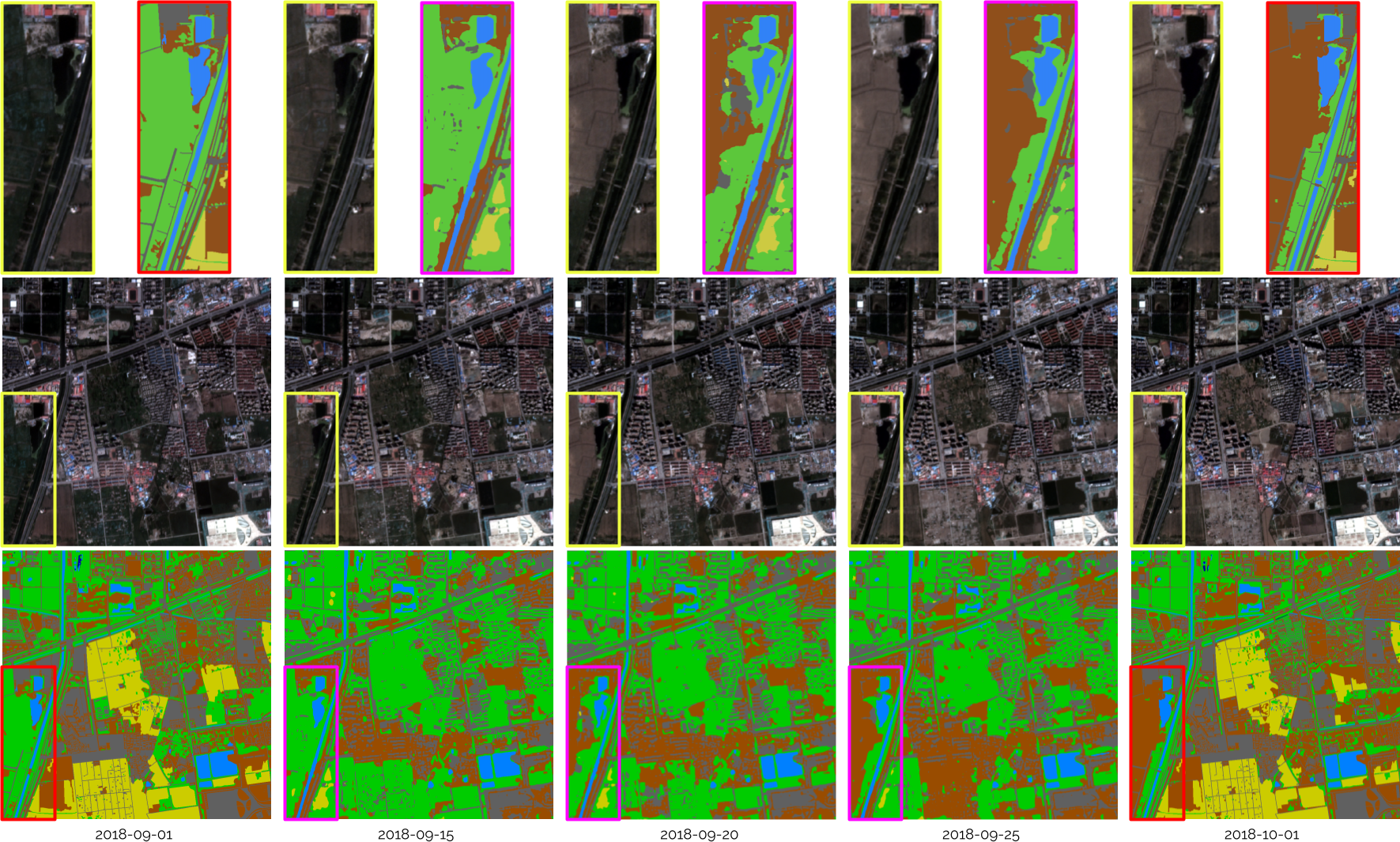}
    \caption{\textbf{Qualitative results on validation set.} 
    Semantic maps (bottom row) of the semi-supervised baseline CAC~\cite{lai2021cvpr} trained on daily images. The input sequence consists of 5 images (middle row) from September to October, spanning one month. For the first and last semantic map of the considered sequence, we show ground-truth labels (bottom right, bottom left). The three middle columns show predictions of~\cite{lai2021cvpr}. For each sample, we magnify a specific area to highlight the temporal transition from forest \& other vegetation to soil, marked red for ground-truth and pink for baseline predictions~\cite{lai2021cvpr}. Notably, this development is captured with high fidelity by our baseline~\cite{lai2021cvpr}. On the other hand, in certain areas, it is not able to distinguish between the generic forest \& vegetation class and the ground-truth label agriculture. For the color representation of segmentation maps see~\cref{table:pixdistribution}.
    } 
    \label{fig:daily_example}
\end{figure*}

\section{Experiments}
\label{sec:baselines}
In this section, we demonstrate the utility of our dataset with various experiments on land cover segmentation and semantic change segmentation. We first give an overview of considered baseline methods in~\cref{subsec:baselines} and then present corresponding results in~\cref{subsec:segmentation} and~\cref{subsec:changesegmentation}.

\subsection{Baselines}
\label{subsec:baselines}
\dynnet\ contains daily images and dense semantic annotations for the first day of each month. This raises the question of how one can leverage additional unlabelled examples to improve the results when training on the labeled data.
We study two separate approaches in this work: spatio-temporal and semi-supervised semantic segmentation. The former addresses the time-series nature of our data by combining spatial information with temporal architectures. The latter uses the annotated images (first day of each month) as supervision while taking advantage of the additional unlabeled samples in an unsupervised manner. 

\PAR{Spatio-temporal baselines.} 
The first class of baselines we consider are spatio-temporal methods. The main idea is to fuse individual observations of an input time series and produce a single output prediction -- the monthly semantic map. As a backbone, we use the U-Net feature extractor~\cite{ronneberger2015u}. Following~\cite{garnot2021iccv,rustowicz2019cvprw}, we compare different temporal
architectures. First, we apply a U-ConvLSTM network~\cite{rustowicz2019cvprw}. As a second method, we utilize 3D convolutions that process spatial and temporal information at once~\cite{rustowicz2019cvprw}. Finally, we employ U-TAE~\cite{garnot2021iccv} that encodes temporal features in the latent space via self-attention~\cite{vaswani2017neurips}.

\PAR{Semi-supervised baselines.} 
As an alternative to modeling the input images as sequences, we can interpret them as an unordered collection of training samples. 
Analogous to standard supervised learning, the labeled examples are used directly as training data. To extract information from the remaining set of unlabeled training examples, we employ the recent state-of-the-art consistency-based semi-supervised segmentation method by Lai \etal~\cite{lai2021cvpr}. The main idea is to randomly crop unlabeled images into pairs of patches and enforce consistent outputs for the overlap of both sub-regions. Robustness to varying contexts is crucial for our data since the surrounding of an overlapping region is generally an unreliable predictor for its class label.
For example, water occurs in quite different environmental contexts in our dataset, like forests, agriculture, or impervious surfaces.
We evaluate this method~\cite{lai2021cvpr} with the segmentation backbone DeepLabv3+~\cite{chen2018eccv}.

\subsection{Land cover and land use segmentation}
\label{subsec:segmentation}
The first task we consider is semantic segmentation of land cover classes. Specifically, the goal is to predict one of the LULC labels described in~\cref{subsec:pixlabels}.
We compare the performance of the two classes of baseline methods discussed in the previous section. 
For each setting, we evaluate the intersection-over-union score averaged over all 6 evaluation LULC classes (mIoU).
Due to its overall scarcity, we exclude the snow \& ice class from the evaluations, see~\cref{subsec:dataprep} for more details. 
\paragraph{Spatio-temporal results.} 

Results of spatio-temporal methods are summarized in~\cref{tab:temporalbaselines}.
As a first reference point, we consider the purely supervised setting. Here, we train a standard U-Net architecture only on the monthly labeled samples. It achieves 33.5\% mIoU on the validation and 37.6\% mIoU on the test set.

We further assess whether existing spatio-temporal architectures benefit from the time-series nature of our data. All three considered architectures improve the performance over the supervised baseline for weekly temporal inputs on the validation set. U-TAE and U-ConvLSTM show the strongest generalization performance on the test set. 

On the other hand, when using daily sequences of 28-31 images, the performance drops considerably. This suggests that generic spatio-temporal techniques are not necessarily optimal for extracting information from daily satellite data. The individual images of such daily time series are often highly correlated. Consequently, when labeled data is limited, increasing the length of a sequence at some point leads to unstable training. For our benchmark, using weekly samples is optimal for the considered baselines. We conclude that more specialized techniques are needed to allow for robust learning on daily time-series satellite imagery.

\paragraph{Semi-supervised results.} 
We report the performances of the baseline~\cite{lai2021cvpr} in combination with DeepLabv3+~\cite{chen2018eccv} in~\cref{tab:semisupervised}.
Similar to the spatio-temporal experiments, we consider different temporal densities. For the purely supervised setting, all unlabeled images are discarded (monthly).
Additionally, we compare different semi-supervised settings with 6 (weekly), 28-31 (daily) unlabelled samples per month. Both, daily and weekly data help to improve over the supervised baseline. 
A detailed analysis of these quantitative results shows that the agriculture and wetland classes prove to be difficult for all baselines. Agricultural areas are often confused with forest or soil, see \cref{fig:daily_example}, whereas wetlands get confused with soil and water. This is, to a certain degree, expected due to the visual similarity of these classes.
Notably, training on daily data achieves the overall best result. The obtained accuracy is 43.6\% mIoU on the test set, with a considerable improvement over the monthly and weekly results of 37.9\%.

\begin{table}[]
\vspace{0.5em}
\begin{center}
\scalebox{0.85}{
\begin{tabular}{llcccc}
\toprule[0.2em]
& &  \textbf{\metric ($\uparrow$)} & \bmetric ($\uparrow$) & \smetric ($\uparrow$) & mIoU ($\uparrow$) \\
\toprule[0.2em]
\parbox[t]{1mm}{\multirow{2}{*}{\rotatebox[origin=c]{90}{\textit{mont.}}}}
  & CAC~\cite{lai2021cvpr}
  & 17.7 & {10.7} & 24.7 & 37.9 \\
  & U-Net~\cite{ronneberger2015u}& 17.3 & 10.1 & 24.4 & 37.6 \\
 \hline
 \hline
 \parbox[t]{1mm}{\multirow{4}{*}{\rotatebox[origin=c]{90}{\textit{weekly}}}}
  & CAC~\cite{lai2021cvpr} & 17.8 & 10.1 & 25.4 & 37.9 \\
 & U-TAE~\cite{garnot2021iccv}&  \textbf{19.1} & {9.5} & \textbf{28.7} & {39.7} \\
 & U-ConvLSTM~\cite{rustowicz2019cvprw} & {19.0} & {10.2} & {27.8} & {39.1} \\
 & 3D-Unet~\cite{rustowicz2019cvprw} & 17.6 & 10.2 & 25.0 & {37.2} \\
\hline
\parbox[t]{1mm}{\multirow{4}{*}{\rotatebox[origin=c]{90}{\textit{daily}}}}
  & CAC~\cite{lai2021cvpr}     & {18.5} & 10.3 & {26.7} & \textbf{43.6} \\
  & U-TAE~\cite{garnot2021iccv}& 17.8 & 10.4 & 25.3 & 36.1 \\
  & U-ConvLSTM~\cite{rustowicz2019cvprw} & 15.6 & 7.0 & 24.2 & 30.9 \\
  & 3D-Unet~\cite{rustowicz2019cvprw} & 18.8 & \textbf{11.5} & {26.1} & 38.8 \\
\bottomrule[0.1em]
\end{tabular}
}
\centering
    \caption{\textbf{Quantitative results of semantic change segmentation on our test set.} This table shows semantic change segmentation results of all methods on our \dynnet\ dataset.}
    \label{tab:changedetection}
\end{center}

\end{table}

\subsection{Semantic change segmentation}
\label{subsec:changesegmentation}
In the following, we compare the performance of our considered baseline methods on the metrics that we introduced in~\cref{sec:change}, see~\cref{tab:changedetection} for results. Similar to~\cref{subsec:segmentation}, we use different degrees of temporal densities with monthly, weekly, and daily observations. As a general trend, the additional weekly observations improve the performance over the purely supervised, monthly baselines. For the semi-supervised approach~\cite{lai2021cvpr} the performance on the test set further improves with daily samples. On the other hand, the benefits from additional daily observations are less consistent for spatio-temporal baselines. In this case, increasing the sequence length is inherently subject to a trade-off between providing more information and decreasing the training stability. Since daily observations are highly correlated, optimal results are achieved for a weekly sampling.

Overall, our results suggest that detecting change (BC) is particularly challenging for our considered baselines. Most obtained accuracies are around $10\%$. Considering that the ground-truth change maps cover only 5\% of all pixels on average, there exist a high number of potential false positives. Oftentimes, change occurs between two classes that are visually very similar, like forest \& other vegetation to soil. The results further confirm that the mIoU metric alone is not sufficient to measure the performance of semantic change segmentation. A high LULC segmentation score (mIoU) does not guarantee optimal performance in terms of the change segmentation score (SCS). When compared directly, the semantic change and binary change performance are somewhat decoupled which warrants the split of our SCS metric into binary change \bmetric\ and semantic change \smetric. 

\section{Conclusion}
\label{sec:conclusion}
We presented DynamicEarthNet, a novel dataset that provides daily, multi-spectral satellite imagery for a broad range of areas of interest. Beyond the raw imagery, it comprises monthly semantic annotations of 7 common LULC classes. This unique combination of dense time-series data and high-quality annotations distinguishes DynamicEarthNet from existing benchmarks, see~\cref{table:relatedsatellites}, which are either temporally sparse or do not provide comparable ground-truth labels. We showed that this gives rise to previously unexplored settings like semi-supervised learning, as well as spatio-temporal methods with an unprecedented temporal resolution. We further devised a new evaluation protocol for semantic change segmentation. It involves several metrics that focus on distinct, common errors in the context of multi-class change prediction. We believe that our benchmark has the potential to spark the development of more specialized techniques that can take full advantage of daily, multi-spectral data. Finally, we highlight in several compelling case-studies how high frequency satellite data can be used to track land cover evolution, \eg due to deforestation, and assess both its short and long-term effects.

\section*{Acknowledgements}
This work is supported by the Humboldt Foundation through the Sofja Kovalevskaja Award, the framework of Helmholtz AI [grant number:  ZT-I-PF-5-01] - Local Unit ``Munich Unit @Aeronautics, Space and Transport (MASTr)", the Helmholtz Association under the joint research school ``Munich School for Data Science - MUDS", and the German Federal Ministry for Economic Affairs and Energy (BMWi) under the grant DynamicEarthNet (grant number: 50EE2005).
%
{\small
\bibliographystyle{plain}
\bibliography{ms}
}
\clearpage

\appendix 
\section{Dataset details}
In the following, we provide additional details on our dataset. For once, we summarize the metadata complementing our raw sensory data in~\cref{sec:planetmetadata}. In~\cref{sec:sentinel2} we describe the auxiliary Sentinel-2~\cite{drusch2012sentinel, aschbacher2017esa} images. Finally, we provide additional information on the different sampling densities of our dataset in~\cref{sec:trainingvis}. 

\subsection{Planet metadata}\label{sec:planetmetadata}
The commercial Planet Fusion data constitutes the core part of the \dynnet\ dataset. In addition to the surface reflectance values (RGB+near-infrared) that we use in the main paper, Planet provides additional quality assurance (QA) information. The purpose of this is to denote which parts of the data are raw observations and which parts are gap-filled with temporally close observations. For every pixel, the QA product gives the distance and direction to the day of the observation. For example, a pixel value of -1 implies that the pixel has been filled from the previous day. 

\subsection{Sentinel 2 auxiliary images}\label{sec:sentinel2}
Sentinel-2 (S2) images are publicly available through the open data policy of the European Space Agency's (ESA) Copernicus Program. The mission collects images of all landmasses every 5 days at a resolution of 10m per pixel~\cite{drusch2012sentinel}. While the temporal and spatial resolution of S2 time-series imagery is smaller than the Planet data, S2 collects 13 channels compared to 4 channels of Planet Fusion. In certain scenarios, the additional channels, particularly in the short-wave infrared spectrum, may provide useful auxiliary information about changes on the ground. 

In order to encourage cross-research between Planet Fusion and S2 data, we accompany our dataset with monthly images of S2 data from the same locations. The Sentinel-2 images are composite images which means they have been created from multiple S2 images throughout the month. This allows for a direct comparison of the effectiveness of different sources of satellite imagery. 

Our Sentinel-2 data is provided as a so-called Bottom-Of-Atmosphere product which includes the correction of distortions to the surface reflectance values caused by atmospheric interference. The S2 pre-processing quality is relatively low compared to the analysis-ready Planet Fusion product. For some areas of interest (AOIs), the collected S2 data suffer from occlusions through cloud coverage for all S2 images in a month. This naturally compromises the quality of the monthly composites. We have collected affected months for all AOIs manually in a designated S2 quality assessment spreadsheet that we provide, together with the dataset. 26\% of monthly S2 composites suffer from minor quality issues and around 5\% have major quality issues. When the community explores applications of S2 data with \dynnet, we advise to investigate whether considered cubes or months are potentially impacted.

\subsection{Temporal densities}\label{sec:trainingvis}
\begin{figure*}
    \centering
    \includegraphics[width=0.9\linewidth]{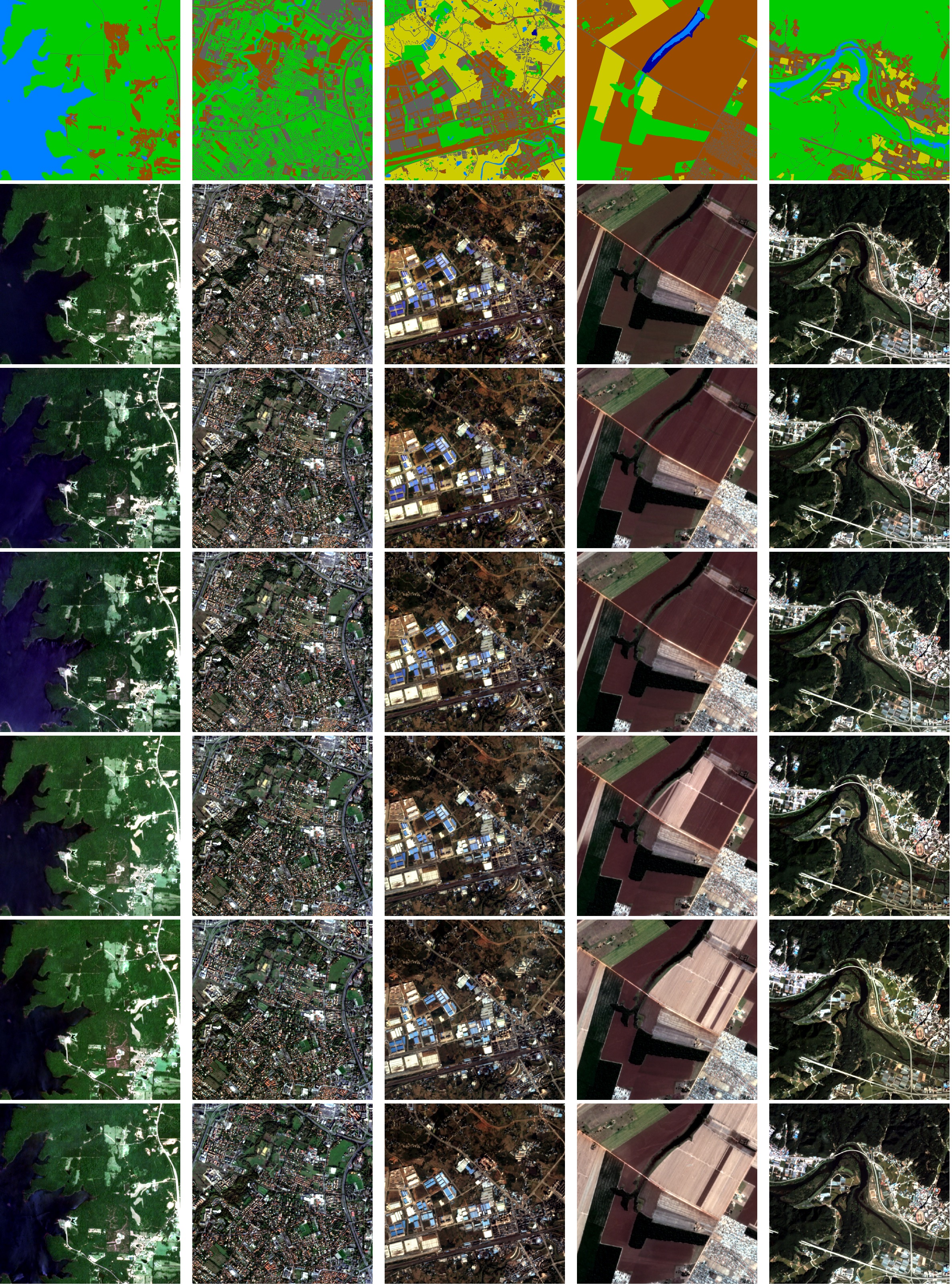}
    \caption{\textbf{Training set samples.} We visualize 5 sample time series (one per column) from the training set of the presented \dynnet\ dataset. Each sequence illustrates weekly samples (row 2-7) and the corresponding annotated monthly labels (1st row).}
    \label{fig:weeklysamples}
\end{figure*}

In our experiments in \cref{subsec:segmentation} and \cref{subsec:changesegmentation}, we use three different temporal sampling densities for both the spatio-temporal and semi-supervised baselines: 
\begin{itemize}
    \item The monthly setting (fully supervised) shows the first day of each month, resulting in a one-to-one correspondence between input images and labels. 
    \item For the weekly setting, we feed the architectures with samples from the 1st, 5th, 10th, 15th, 20th and 25th days of each month.  
    \item The daily setting uses all the available images in a considered month, as well as the corresponding monthly label.
\end{itemize}
In~\cref{fig:weeklysamples}, we show the images of 5 time-series with a weekly sampling density. 

\section{Evaluation protocol details}
In the following, we motivate our design choices for the metric proposed in \cref{sec:change} and compare it to other existing metrics.

\subsection{Semantic change}
In contrast to semantic segmentation, semantic change segmentation focuses on the changed parts of a given semantic map. Similar to how boundary segmentation restricts evaluation to the boundary pixels, our proposed metric is restricted to changed pixels. We consider several options on how to restrict this subset. In the following, we refer to pixels that have changed their semantic class from one timestep to the next as changed pixels:

\begin{itemize}
    \item[\textbf{R1.}] We restrict the evaluation to the set of changed pixels, as predicted by the considered method.
    \item[\textbf{R2.}] We restrict the evaluation to the set of changed pixels defined by the ground-truth semantic maps.
    \item[\textbf{R3.}] We restrict the evaluation to the intersection of R1 and R2, which is the set of true positives.
\end{itemize}

Using the set of R1 or R3 has the disadvantage that it couples the semantic change performance with the binary change performance. Only the pixels that are predicted as changed are potentially also evaluated for the semantic change score. Hence, errors in the binary change influence the semantic change score, which potentially opens the metric to misconduct. One can easily imagine a method that reduces the set of predicted change artificially to a single pixel for which the semantic class is predicted with very high confidence. Then the \smetric\ score would be perfect (1.0), while the \bmetric\ score would be close to 0. The resulting overall \metric score would be around 0.5, which is much higher than the scores reported in~\cref{tab:changedetection}. Such behavior is completely undesired and leads to a metric that is not aligned with human intuition, with results that are hard to interpret. Thus, we use the second option R2 to compute the \smetric\ metric. This makes the errors decoupled and the scores easy and intuitive to interpret.

\subsection{Comparison}
Even though there exists no unified evaluation protocol for semantic change segmentation, there are a few metrics that focus on certain aspects of the task. In the following, we discuss the different options and compare their efficacy for the task of semantic change segmentation.

\PAR{Pixel accuracy.} Pixel accuracy, also referred to as overall accuracy, is one of the simplest measures for (binary) segmentation problems. It is defined as the ratio of correctly classified pixels to all pixels. In settings like ours, in which there are 2 classes for binary change and a high imbalance between them, the pixel accuracy is not able to report meaningful insights. In our setting, 95\% of all pixels do not change. Thus, a score of 95\% can be obtained by predicting no change all the time. Therefore, we refrain from using pixel accuracy as a metric.

\PAR{mIoU.} The standard mean intersection-over-union addresses the immediate shortcomings of the vanilla pixel accuracy metric. It is possible to use it for both, binary change and semantic change. However, using the mIoU metric for binary change directly, \ie computing the mean IoU of the 2 classes, suffers also from the imbalance issues discussed for the pixel accuracy. Thus, the proposed \bmetric\ metric computes the IoU of only the change class, rather than both the change and no-change class. For the semantic change, we however apply mIoU, \ie computing the mean over all semantic classes. As explained in the previous subsection, an insightful change metric should focus on the changed regions. We, therefore, refrain from using mIoU on the whole image but compute the scores solely on the changed pixels.

\PAR{Cohen's kappa.} Previous works~\cite{lyu2016remotesensing,mou2019spectral,tian2020neuripsw} have used Cohen's kappa to measure the performance in similar settings. Cohen's kappa is a statistical measure of the agreement between the predictions and ground-truth. It is more robust compared to pixel accuracy as it takes the agreement occurring by pure chance into account. However, this measure is not as informative as mIoU. It does not offer insights into the performance of individual classes. Moreover, since scores are not aggregated per class, the performance of classes with high appearance rates will dominate the score and therefore lead to an overall higher score. For more details about the dataset imbalance, we refer to \cref{table:pixdistribution}. We thus choose to adapt the well-established IoU measure for our needs.

\subsection{Correcting wrong predictions}
Our proposed metric requires a separate binary change map $\tbh$ and semantic map $\tyh$. It is therefore not limited to the special case of computing the binary change $\tbh$ directly from the predicted semantic maps for two consecutive timesteps $\tyh_{t-1}$ and $\tyh_{t}$. This provides additional flexibility, as it is often preferable to decouple the semantic maps from the change predictions~\cite{lyu2016remotesensing,mou2019spectral}. Moreover, it allows for the correction of previous mistakes in online methods that obtain predictions frame-by-frame for an input time-series. As an example, suppose that a semantic class for a certain pixel is predicted wrong at a given timestep. If that pixel does not change in the next timestep, its prediction would either need to keep the wrong semantic class or predict a different semantic class. However, predicting a different semantic class would automatically be recognized as a predicted change, resulting in an error in the binary change. Thus, there is no way to correct previous mistakes without introducing another one. This also holds for other types of errors. By requiring each method to pass explicitly a binary change map $\tbh$ and semantic map $\tyh$, this issue can be avoided. In our setting and the above example, the semantic class can be corrected without predicting a binary change for this pixel. This is especially important for methods that are used for both semantic segmentation as well as semantic change segmentation.

\subsection{Discussion on bi-temporal change}
\begin{table}[]
\vspace{0.5em}
\begin{center}
\scalebox{0.85}{
\begin{tabular}{llccc}
\toprule[0.2em]
& &  \textbf{\metric ($\uparrow$)} & \bmetric ($\uparrow$) & \smetric ($\uparrow$) \\
\toprule[0.2em]
\parbox[t]{1mm}{\multirow{4}{*}{\rotatebox[origin=c]{90}{\textit{bi-temp}}}}
  & CAC~\cite{lai2021cvpr} & 17.8 & 10.1 & 25.4 \\
 & U-TAE~\cite{garnot2021iccv}& \textbf{19.1} & {9.5} & \textbf{28.7} \\
 & U-ConvLSTM~\cite{rustowicz2019cvprw} & {19.0} & \textbf{10.2} & {27.8} \\
 & 3D-Unet~\cite{rustowicz2019cvprw} & 17.6 & 10.2 & 25.0  \\
\hline
\hline
\parbox[t]{1mm}{\multirow{4}{*}{\rotatebox[origin=c]{90}{\textit{multi-temp}}}}
  & CAC~\cite{lai2021cvpr} & \textbf{27.7} & 23.6 & \textbf{31.8}  \\
 & U-TAE~\cite{garnot2021iccv}&  27.6 & {23.4} & 31.8  \\
 & U-ConvLSTM~\cite{rustowicz2019cvprw} & {27.5} & \textbf{24.2} & {30.7} \\
 & 3D-Unet~\cite{rustowicz2019cvprw} & 25.3 & 21.2 & 29.4  \\

\bottomrule[0.1em]
\end{tabular}
}
\centering
    \caption{\textbf{Quantitative results of our metric variant on our test set.} The first row shows the bi-temporal, and the second row shows the multi-temporal results on weekly data. The first row results are identical to the weekly results in~\cref{tab:changedetection}.}
    \label{tab:bi-multitemporal}
\end{center}
\end{table}

In~\cref{subsec:problemdef}, we define the problem as a bi-temporal semantic change segmentation that measures the SCS, SC, and BC scores for a given ground truth $\ty_{t}$ and $\ty_{t+1}$. Given that our dataset contains consistent multi-temporal land use and land cover ground-truth information, it allows us to extend the bi-temporal metric definition and calculate the scores on time intervals of varying lengths. Specifically, we investigate a variant of our bi-temporal semantic change segmentation metrics by measuring the change between all viable pairs of months ($t$ to $t+1$, $t+2$, $t+3$, ....) at each area of interest (24 x 23 = 552 pairs in total).

We report the resulting accuracies in~\cref{tab:bi-multitemporal}.
For the most part, the modified metric yields slightly higher values than our bi-temporal metric. We attribute this to the fact that the modified metric has a certain smoothing effect, \ie less emphasis is placed on pinpointing the exact frame where change occurs. Throughout our dataset, we notice that different types of changes happen over different time periods (daily, weekly, monthly quarterly, or even seasonally/yearly). On the other hand, the smoothing effect of longer time intervals potentially under-penalizes prediction errors on small time intervals, which goes against one of the main motivations of having daily time-series observations. In our work, we ultimately prefer the bi-temporal setting and leave the detailed multi-temporal discussion as future work.

\section{Implementation details}
All the experiments are implemented in PyTorch. Our dataset contains 4 spectral bands (RGB + near-infrared). The theoretical valid range for all 4 channels is 1-32,767; however, in practice, the maximum value for the type of data contained in our dataset is 10,000.  For data normalization, we calculate the mean and standard deviation per band, averaged over the whole dataset. The exact obtained values are
\begin{align*}
\mathrm{mean}&=[1042.59, 915.62, 671.26, 2605.21]\qquad\text{and}\\\mathrm{std}&=[957.96, 715.55, 596.94, 1059.90],
\end{align*} 
respectively. For data augmentation, we randomly resize the images with a ratio between $[0.5, 2]$ and crop them to half the input resolution $(512, 512)$. Additionally, we apply random horizontal flips. As we specified in~\cref{subsec:dataprep}, due to the scarcity of the snow \& ice class, we do not include them in the test and validation set. For the spatio-temporal architectures, we use the Adam optimizer with a learning rate of $1e-4$. The batch size is set to 4. We generally train our networks for up to 100 epochs. For the spatio-temporal experiment with daily samples, we use 200 epochs to ensure convergence. The reported results are taken from the epoch that achieves the highest validation accuracy. For the semi-supervised architecture, analogous to~\cite{lai2021cvpr}, we use the SGD optimizer with the poly learning rate decay policy. For both the supervised and unsupervised samples, we use a batch size of 8.

\section{Additional qualitative results}
\begin{figure*}
    \centering
    \vspace{1em}
    \begin{overpic}
    [width=0.9\linewidth]{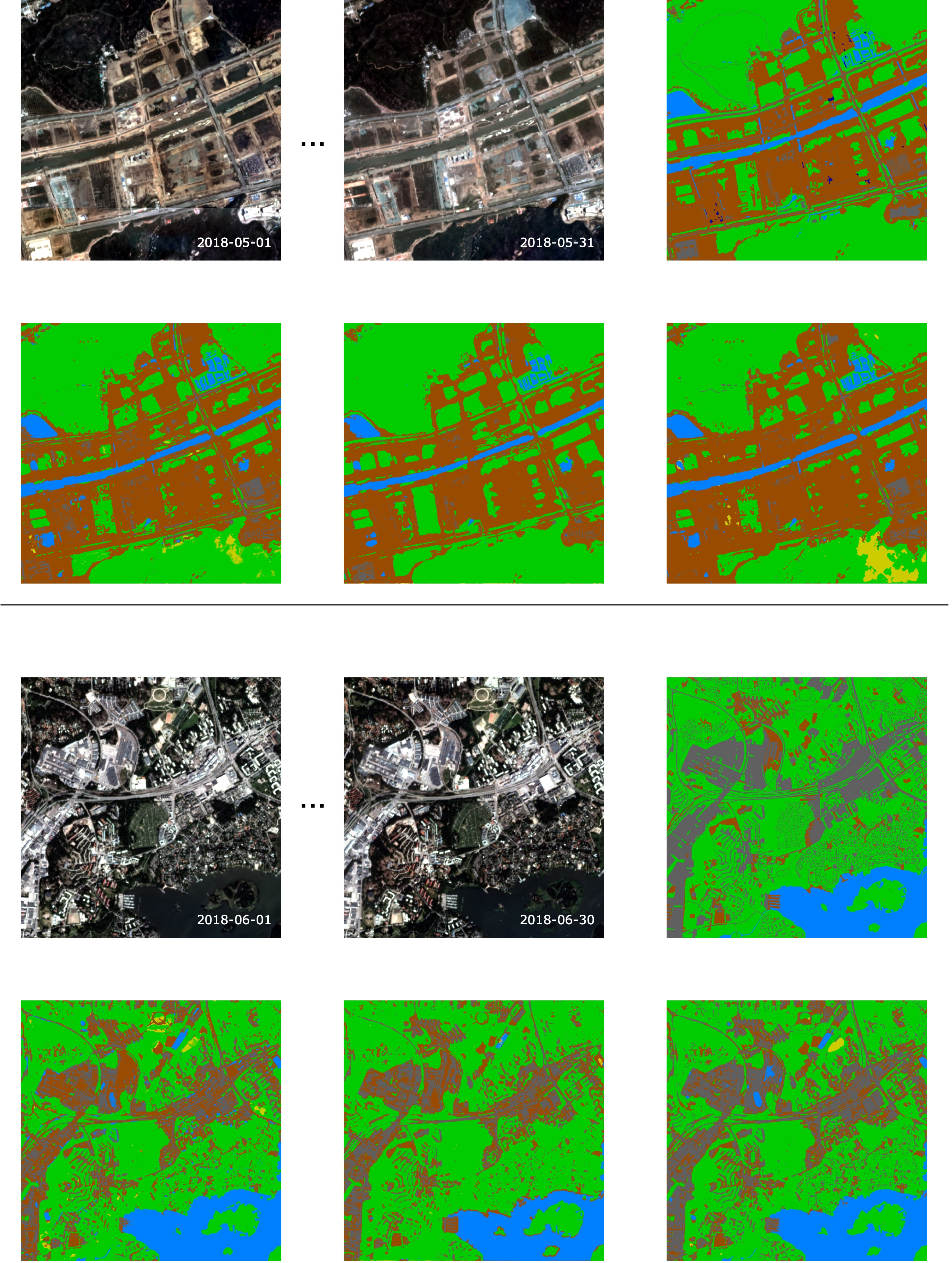}
    \put(19,101.5){Input time-series}
    \put(56.5,101.5){Target semantic map}
    \put(8.5,76){U-TAE~\cite{garnot2021iccv}}
    \put(32,76){U-ConvLSTM~\cite{rustowicz2019cvprw}}
    \put(59,76){3D-Unet~\cite{rustowicz2019cvprw}}
    \put(19,48){Input time-series}
    \put(56.5,48){Target semantic map}
    \put(8.5,22.25){U-TAE~\cite{garnot2021iccv}}
    \put(32,22.25){U-ConvLSTM~\cite{rustowicz2019cvprw}}
    \put(59,22.25){3D-Unet~\cite{rustowicz2019cvprw}}
    \end{overpic}
    \caption{\textbf{Spatio-temporal predictions.} We show two qualitative comparisons of the spatio-temporal methods discussed in \cref{subsec:baselines}. Both examples are taken from our validation set. The methods take a sequence of 31 and 30 daily samples as inputs (top left) and predict a single semantic map for the whole month (bottom row). We furthermore show the ground-truth annotated map for comparison (top right).}
    \label{fig:spatiotemporalsamples}
\end{figure*}

\begin{figure*}
    \centering
    \begin{overpic}
    [width=0.9\linewidth]{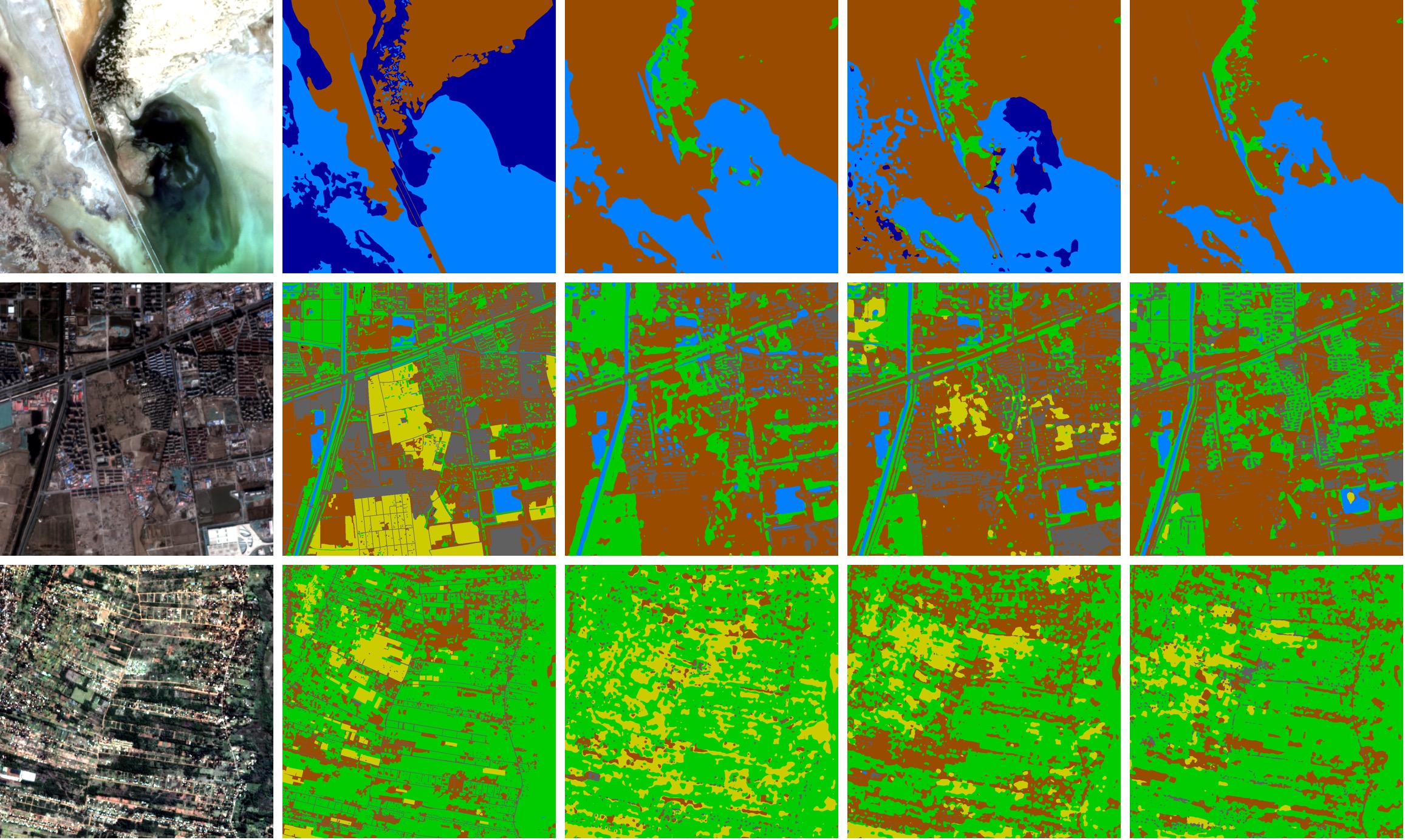}
    \put(6,61){Input}
    \put(26,61){Target}
    \put(46,61){Monthly}
    \put(66,61){Weekly}
    \put(86,61){Daily}
    \end{overpic}
    \caption{\textbf{CAC~\cite{lai2021cvpr} predictions.} We show sample predictions by the semi-supervised baseline CAC~\cite{lai2021cvpr} for three different examples from our validation set. For each example, we depict the input sample (1st column), the ground-truth semantic map (2nd column), as well as the predictions of~\cite{lai2021cvpr} for the monthly, weekly, and daily training setup (3rd-5th column) respectively.}
    \label{fig:deeplabtemporal}
\end{figure*}

\begin{figure*}[h!]
    \centering
    \begin{overpic}
    [width=0.47\linewidth]{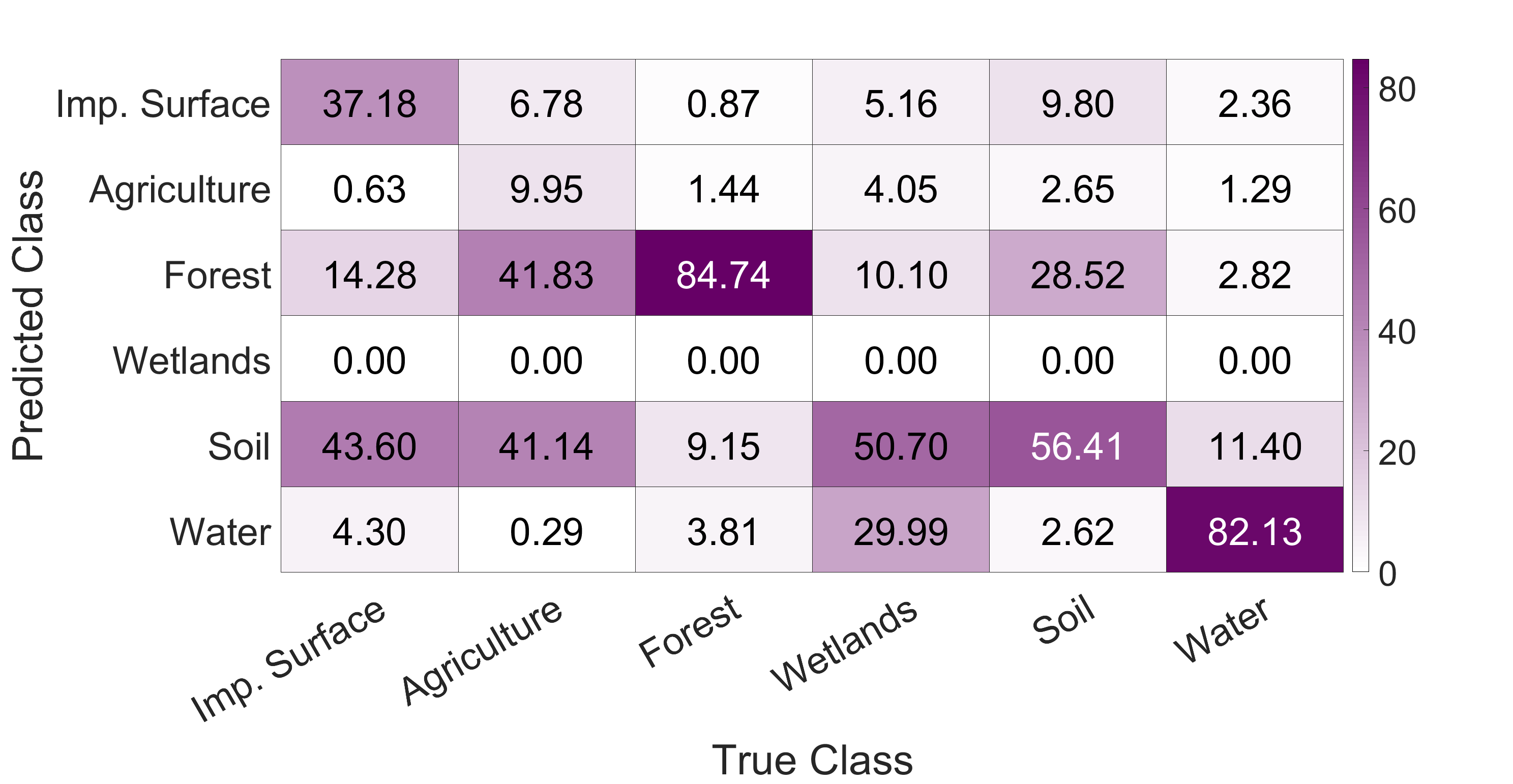}
    \put(30,55){Daily U-TAE~\cite{garnot2021iccv}}
    \end{overpic}
    \begin{overpic}
    [width=0.47\linewidth]{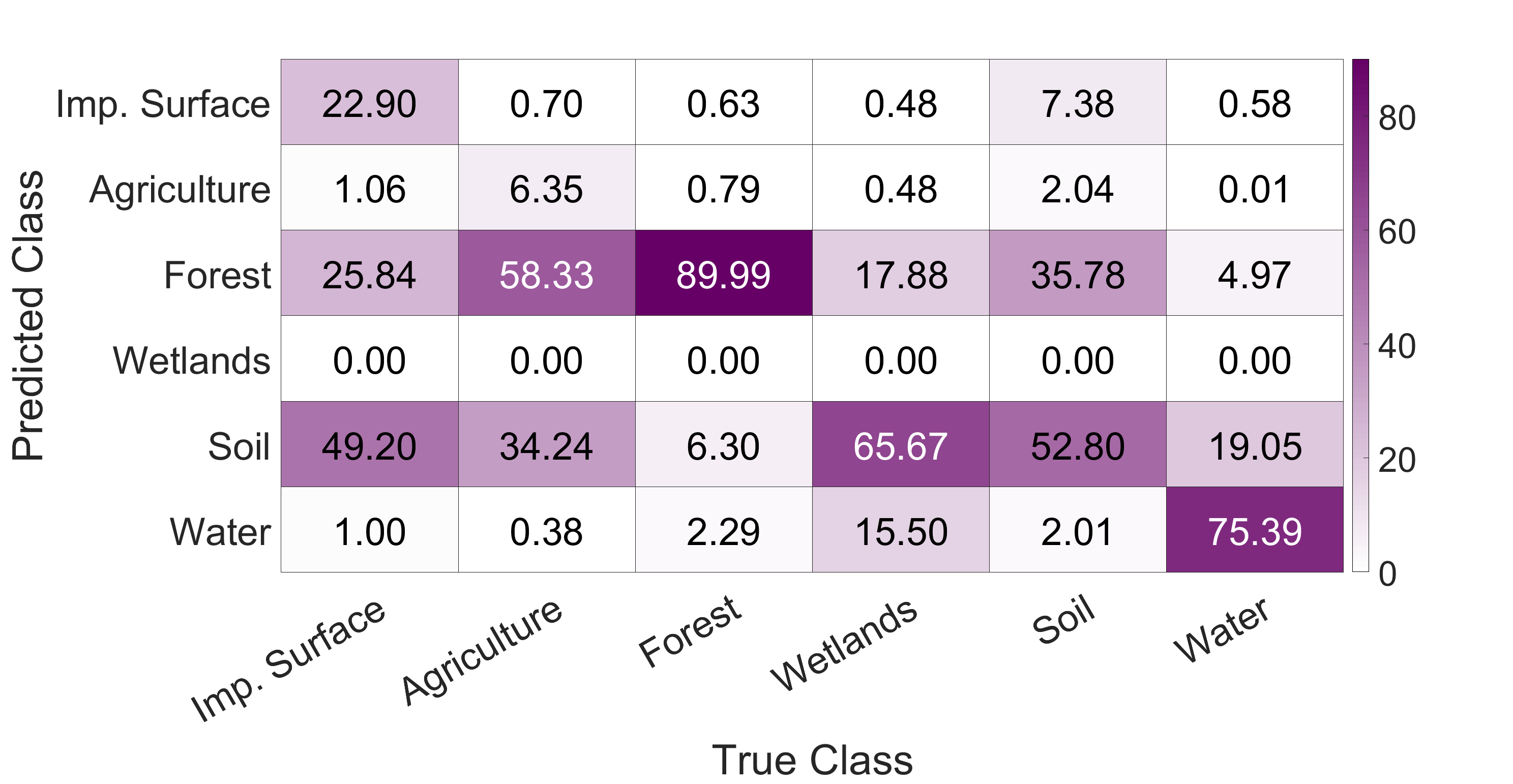}
    \put(33,55){Monthly CAC~\cite{lai2021cvpr}}
    \end{overpic}
    \vspace{3em}
    
    \begin{overpic}
    [width=0.47\linewidth]{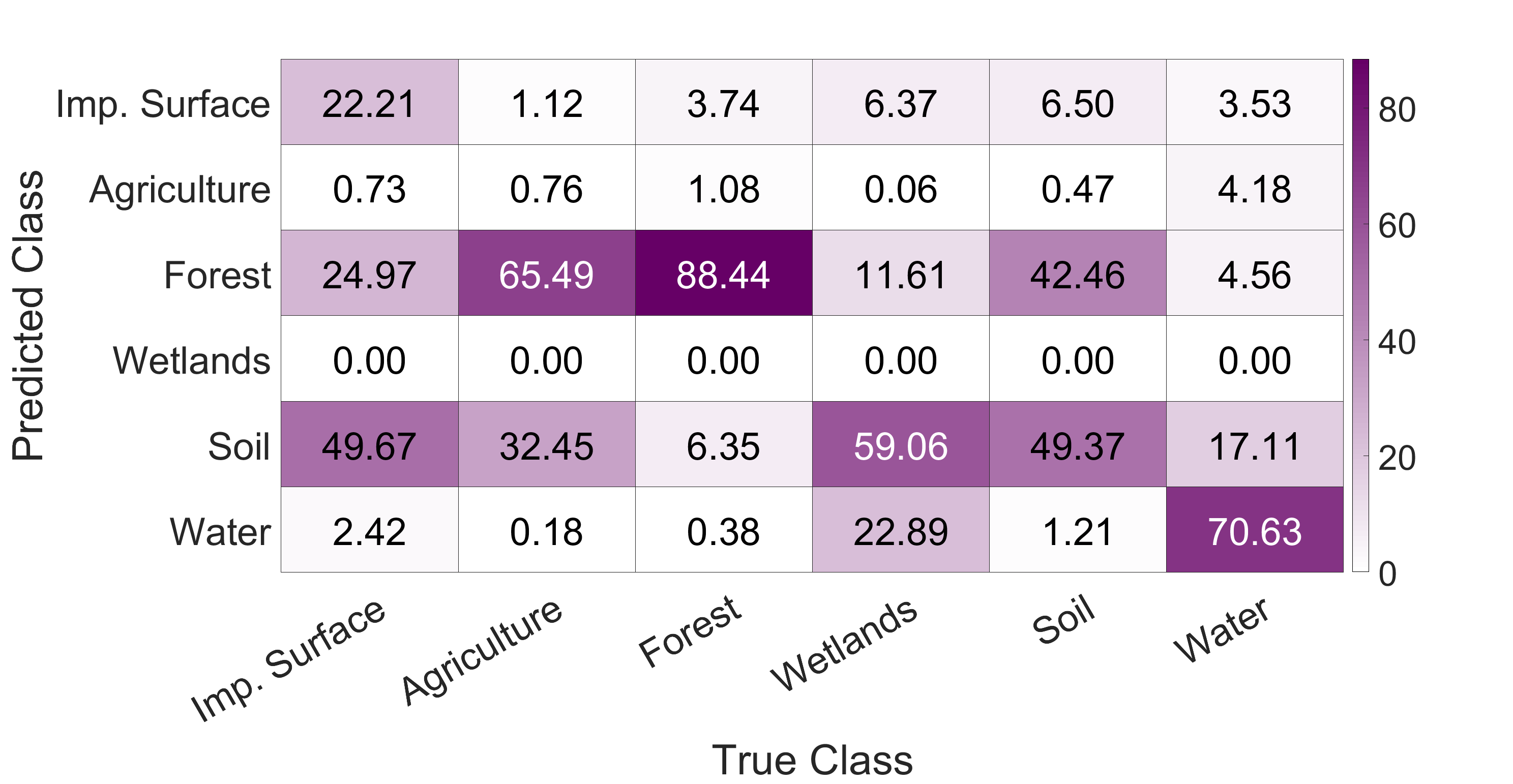}
    \put(30,55){Daily U-ConvLSTM~\cite{rustowicz2019cvprw}}
    \end{overpic}
    \begin{overpic}
    [width=0.47\linewidth]{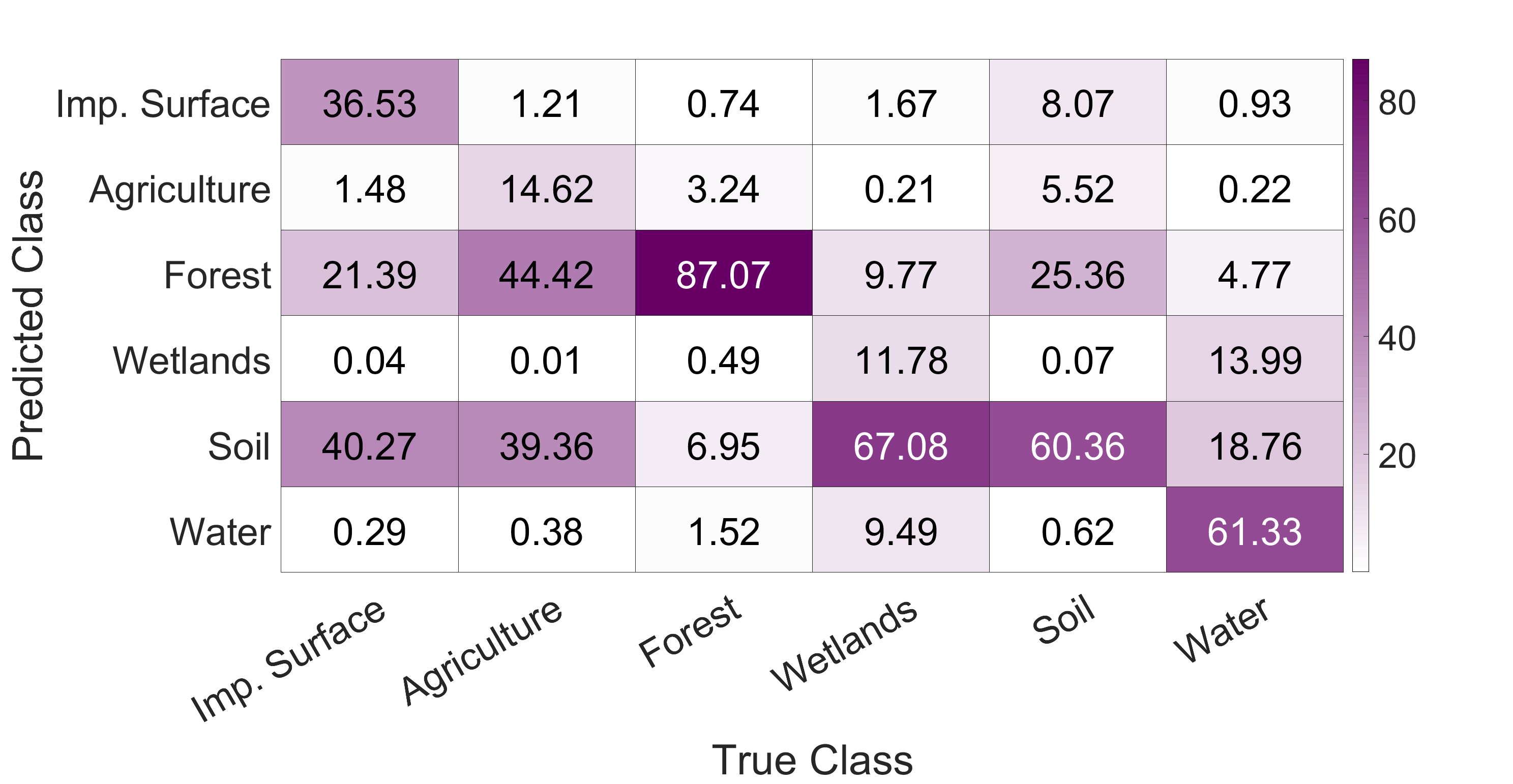}
    \put(33,55){Weekly CAC~\cite{lai2021cvpr}}
    \end{overpic}
    \vspace{3em}
    
    \begin{overpic}
    [width=0.47\linewidth]{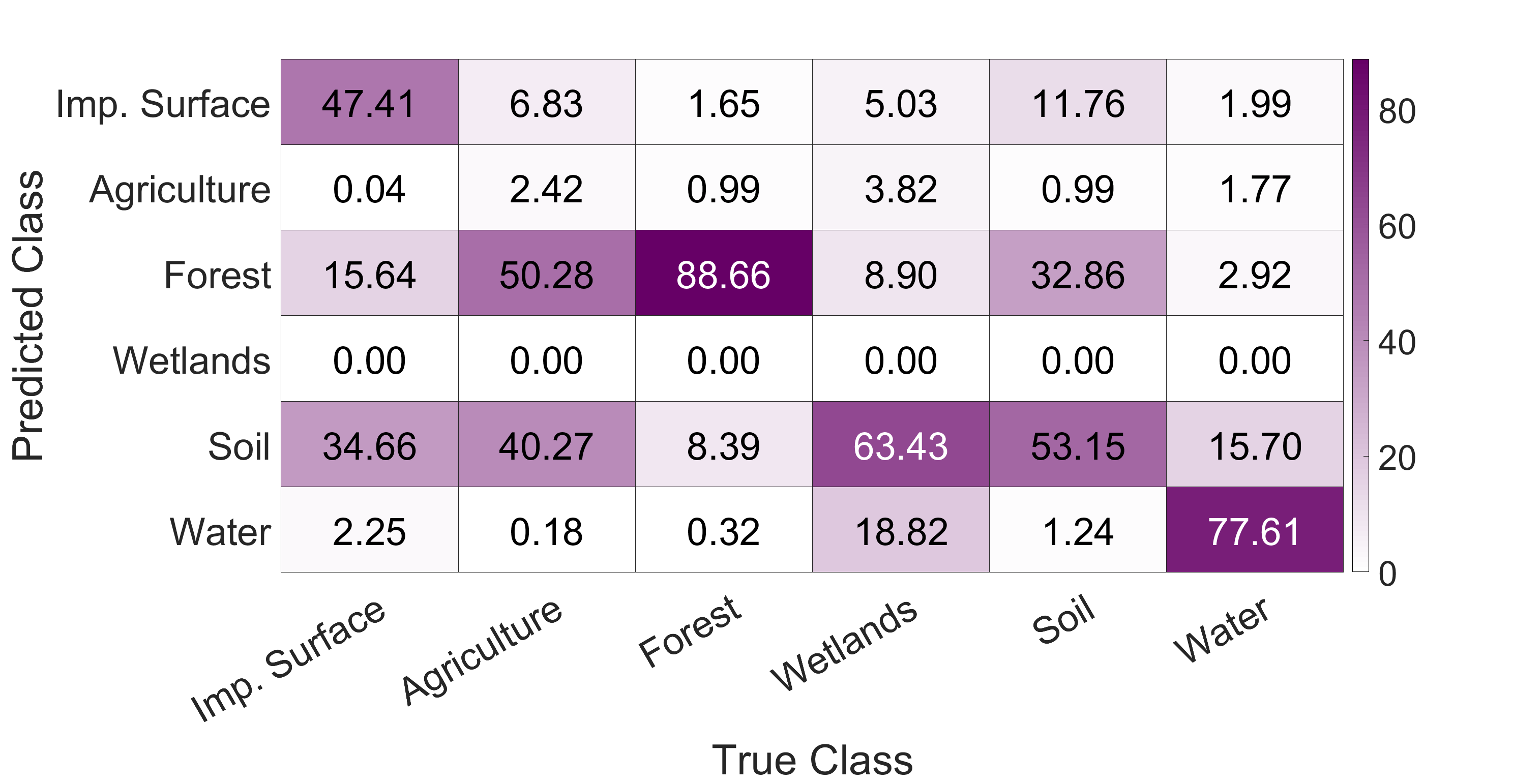}
    \put(30,55){Daily 3D-Unet~\cite{rustowicz2019cvprw}}
    \end{overpic}
    \begin{overpic}
    [width=0.47\linewidth]{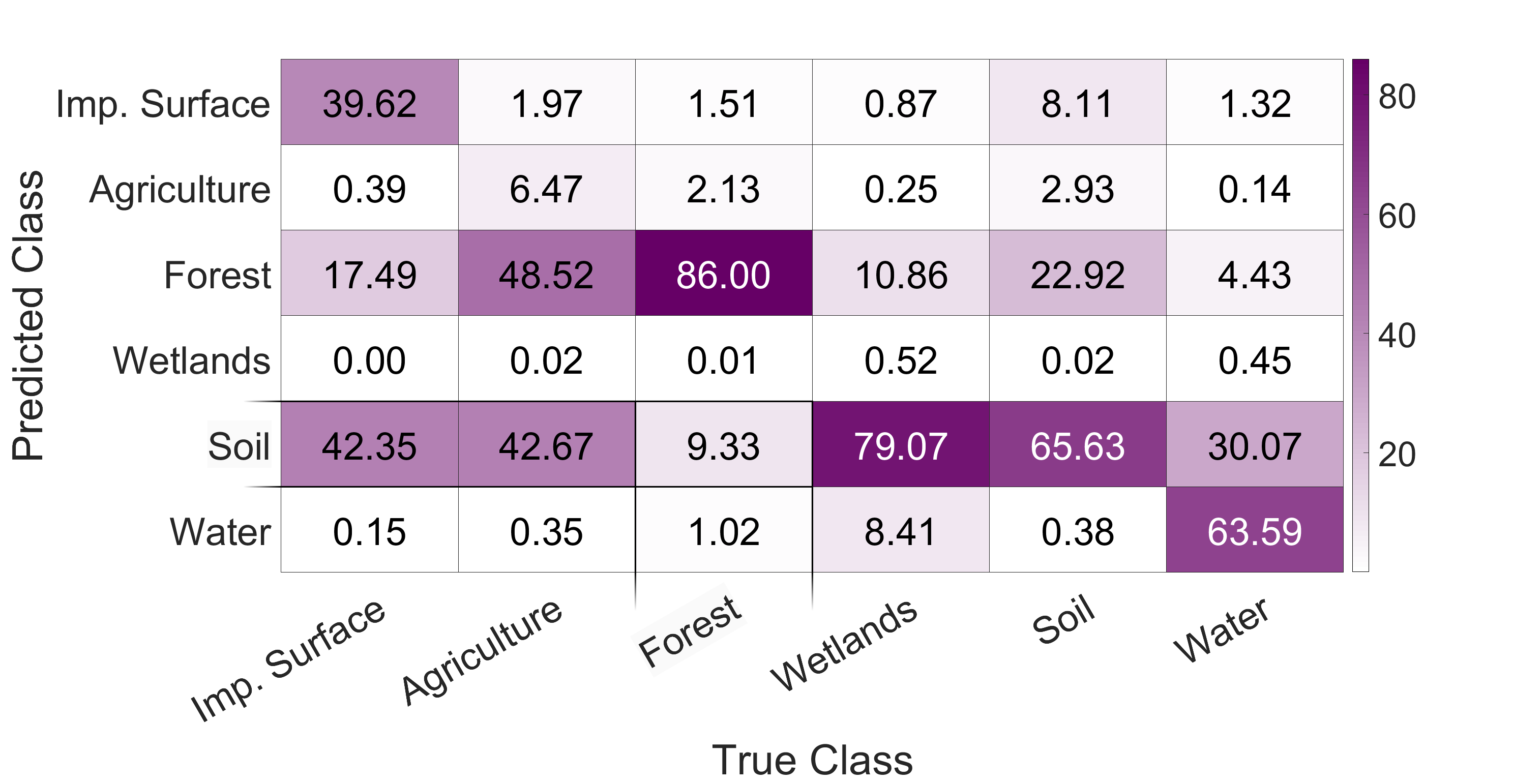}
    \put(33,55){Daily CAC~\cite{lai2021cvpr}}
    \end{overpic}
    \caption{\textbf{Confusion matrices.} We show confusion matrices corresponding to the LULC segmentation results in \cref{subsec:segmentation} on the validation set. The goal is to provide a fine-grained analysis of which classes frequently get misclassified as certain other classes. Each column of an individual confusion matrix is normalized, meaning that it shows the relative distribution of predictions (in percent) for a given, true class. Results are shown for both spatio-temporal (left column) and semi-supervised baselines (right column) with three different settings each.}
    \label{fig:confmatrix}
\end{figure*}

\PAR{Additional visualizations.}
We present additional qualitative visualizations corresponding to the results in \cref{subsec:segmentation}. In \cref{fig:spatiotemporalsamples}, we depict a comparison of the different spatio-temporal baselines described in~\cref{subsec:baselines}.
Furthermore, we compare the effect of different temporal densities for the semi-supervised baseline CAC~\cite{lai2021cvpr} in~\cref{fig:deeplabtemporal}. The weekly training achieves the best results on the validation set, as indicated by the results in~\cref{tab:semisupervised}. This is mostly due to the fact that monthly and daily settings struggle to predict uncommon classes like wetlands (first example in~\cref{fig:deeplabtemporal}) and agriculture (second example in~\cref{fig:deeplabtemporal}). Note that these observations are consistent with the confusion matrices shown in~\cref{fig:confmatrix}.

\PAR{Confusion matrices.}
We provide confusion matrices to complement our results on LULC segmentation in \cref{subsec:segmentation}. The main idea is to allow for a more fine-grained analysis in terms of the 6 semantic classes, see \cref{fig:confmatrix}. We show results for both spatio-temporal methods and semi-supervised learning. Each confusion matrix depicts which classes typically get misclassified as certain other classes. For example, the overall uncommon class wetland frequently gets mislabeled as soil, see \eg the first example in \cref{fig:deeplabtemporal}. Beyond that, one can also directly read the relative segmentation accuracy of each class in the diagonal entries. As can be expected, the predictions are overall more stable for the more common classes like forest \& other vegetation and soil, see \cref{table:pixdistribution} for reference. Among the spatio-temporal methods, the 3D-Unet~\cite{rustowicz2019cvprw} setting yields the best results for the challenging impervious surface class, see \eg the first example in \cref{fig:spatiotemporalsamples}. All in all, these results indicate that future approaches might benefit from reweighting the individual class labels for a more balanced training that can account for rare LULC classes. 

\end{document}